\documentclass{article}

\usepackage[arxiv, nonatbib]{neurips_2023}

\usepackage[utf8]{inputenc} % allow utf-8 input
\usepackage[T1]{fontenc}    % use 8-bit T1 fonts
\usepackage{hyperref}       % hyperlinks
\usepackage{url}            % simple URL typesetting
\usepackage{booktabs}       % professional-quality tables
\usepackage{amsfonts}       % blackboard math symbols
\usepackage{nicefrac}       % compact symbols for 1/2, etc.
\usepackage{microtype}      % microtypography
\usepackage{xcolor}         % colors
\usepackage{amsmath}
\usepackage{amssymb}
\usepackage{multirow}
\usepackage[pdftex]{graphicx}
\usepackage{caption} 
\usepackage[numbers]{natbib}
\title{INT2.1: Towards Fine-Tunable Quantized Large Language Models with Error Correction \\through Low-Rank Adaptation}
\author{%
  Yuji Chai\thanks{These authors contributed equally.}\hspace{2mm}$^{1,3}$, John Gkountouras\footnotemark[1]\hspace{2mm}$^{2,3}$, Glenn G. Ko$^{1,3}$, David Brooks$^{1}$, Gu-Yeon Wei$^{1}$\\
  $^1$Harvard University, $^2$University of Amsterdam, $^3$Stochastic Inc.\\
  \texttt{yuc927@g.harvard.edu, john.gkountouras@student.uva.nl}\\
}

\begin{document}

\maketitle

\begin{abstract}
    We introduce a method that dramatically reduces fine-tuning VRAM requirements and rectifies quantization errors in quantized Large Language Models. First, we develop an extremely memory-efficient fine-tuning (EMEF) method for quantized models using Low-Rank Adaptation (LoRA), and drawing upon it, we construct an error-correcting algorithm designed to minimize errors induced by the quantization process. Our method reduces the memory requirements by up to $5.6\times$, which enables fine-tuning a $7$ billion parameter Large Language Model (LLM) on consumer laptops. At the same time, we propose a Low-Rank Error Correction (LREC) method that exploits the added LoRA layers to ameliorate the gap between the quantized model and its float point counterpart. Our error correction framework leads to a \textbf{fully functional INT2} quantized LLM with the capacity to generate coherent English text. To the best of our knowledge, this is the first INT2 Large Language Model that has been able to reach such a performance. The overhead of our method is merely a $1.05\times$ increase in model size, which translates to an effective precision of \textbf{INT2.1}. Also, our method readily generalizes to other quantization standards, such as INT3, INT4, and INT8, restoring their lost performance, which marks a significant milestone in the field of model quantization. The strategies delineated in this paper hold promising implications for the future development and optimization of quantized models, marking a pivotal shift in the landscape of low-resource machine learning computations. To aid the reproducibility and adoption of our work, we open-source our implementation \href{https://github.com/stochasticai/xturing/blob/main/examples/int4_finetuning/README.md}{here}.
    
    %%The primary objective of these methods is to improve the accuracy of extremely low-bit quantized Large Language Models, specifically targeting the challenges associated with \textbf{INT2} quantization. 
\end{abstract}

\section{Introduction}
Large Language Models (LLMs) of the transformer architecture \cite{Vaswani} demonstrate significant potential in the field of Natural Language Processing (NLP) \cite{GPT2, PALM}. Their proficiency emerges from training on extensive datasets, often encompassing trillions of words, which equips them with a profound understanding of linguistic patterns, grammar, and contextual relationships. This learning process exposes the model to a myriad of textual modalities, ranging from scientific papers to social media posts, thus facilitating a broad spectrum of linguistic style and domain knowledge acquisition.

Despite their impressive capabilities, the optimal performance of LLMs cannot be achieved through the mere utilization of pre-trained models, especially in domain-specific applications. LLMs are generally trained on datasets that contain broad information from the internet.\cite{PALM, GPT3} Consequently, they lack specific knowledge pertinent to specialized fields such as biology, finance, or health care. Projects like BioGPT \cite{bioGPT} and BloombergGPT \cite{BloombergGPT} have underscored the efficacy of fine-tuning in enhancing performance and outpacing general-purpose models such as GPT-3 \cite{GPT3}.
%Frequent updates are essential for LLMs to remain current, similar to the requirements of a recommender system\cite{something}.

Nonetheless, the remarkable capabilities of LLMs come at a cost; they necessitate significant computational resources for fine-tuning and deployment. According to \cite{wei2022emergent}, certain emergent abilities of LLMs, such as reasoning, only appear once the LLM reaches a particular scale. The large size of these models is crucial to their capacity for generating contextually relevant responses and engaging in conversations that mimic human-like interactions. However, this also implies a more substantial memory requirement, which can be prohibitive for enterprises or personal users aiming to fine-tune LLMs for their specific needs.

Previous efforts aimed at reducing memory cost for model fine-tuning have largely centered around two research areas: quantization of neural networks and parameter-efficient fine-tuning. Quantization reduces the number of bits required to represent each parameter in the model. GPTQ\cite{GPTQ}, a prominent method in this area, has successfully compressed LLMs down to 4 bits or lower, albeit at the expense of some loss in downstream performance. However, its quantization methodology doesn't enable lower memory usage during LLM fine-tuning due to its design as an inference solution.

Parameter-efficient fine-tuning (PEFT) methods reconceptualize the fine-tuning process. Given that fine-tuning LLMs is essentially a transfer learning process with only minor revisions to the parameters, PEFT methods exploit this by introducing or selecting a small fraction of parameters compared to the baseline model. One notable example of this approach is Low-Rank Adaptation of LLMs (LoRA)\cite{LORA}, which injects a minimal number of additional parameters in the form of low-rank decomposition matrices into the LLM, thereby enabling efficient fine-tuning.

Our work draws inspiration from LoRA and integrates it into a novel memory-efficient fine-tuning framework. We introduce an error correction scheme for quantized models that is agnostic to the quantization method. This framework significantly reduces memory requirements, thus rendering the fine-tuning of LLMs feasible on devices with limited resources, such as personal laptops. We present the following key contributions:

\begin{enumerate}
\item An Extremely Memory-Efficient Finetuning (EMEF) framework that integrates low-rank adaptation, reducing memory requirement by 5.6x and enabling fine-tuning of LLMs on lower-resource computing devices, such as a consumer-grade laptop. 
\item A quantization-agnostic error correction framework, Low-Rank Error Correction (LREC), that exploits additional floating-point parameters inserted for fine-tuning to mitigate downstream performance loss due to quantization, outperforming quantization baselines.
\item A fully functional INT2 Large Language Model that is capable of generating coherent human-level text, outperforming models compressed using prior techniques.
\end{enumerate}
\section{Background}

\subsection{Large Language Models}

The field of natural language processing (NLP) has undergone significant progress in recent years, notably due to the development of large language models. These models, composed of billions of parameters, are trained on extensive volumes of text data, facilitating an understanding and generation of human-like language \cite{GPT2, GPT3, OPT, UL2}. These models are adept at capturing the intricacies and subtleties of human language, thereby enabling a myriad of language-related tasks such as --inter alia-- text generation, translation, summarization, and question answering, and code generation \cite{flan-T5, codegeex}

In this study, we focus on the Generative Pre-trained Transformer (GPT) models. These models have exhibited remarkable performance in language modeling tasks, and they form the backbone of popular AI applications such as ChatGPT \cite{GPT4}.

\subsection{Quantization for Deep Networks}

Quantization serves as a valuable approach for reducing the size of deep networks. This method diverges from the standard 32-bit floating-point representation for each parameter, instead employing a fewer number of bits for each parameter in the compressed model. Widely adopted quantization standards include FP16 and INT8, which utilize 16-bit floating point and 8-bit integer number formats respectively \cite{llmint8, int8, nuqmm}. However, either the reduction in size is often accompanied by a trade-off in accuracy or the solution is inference-only. Due to the reduced bit usage for parameters, the model's output starts to deviate from its original floating-point baseline. Notably, the methods developed in \cite{brecq} tackle most of these issues, but the solution is not easily scalable to the massive amounts of parameters in Large Language Models.
% A number of strategies, such as quantization smoothing and quantization-aware training, have been proposed to offset this accuracy loss \cite{smoothquant, QAT}.

A more recent approach, GPTQ\cite{GPTQ}, has emerged as a state-of-the-art method for quantization. GPTQ is a post-training quantization method that requires no retraining or access to the full training dataset, yet significantly enhances the accuracy of models compressed to $4$ bits or lower. GPTQ employs a one-shot weight quantization method that uses approximate second-order (Hessian) information to achieve high accuracy and efficiency. It can quantize decoder-only transformer models with $175$ billion parameters in about four GPU hours, reducing the bandwidth to $3$ or $4$ bits per weight, whilst maintaining accuracy levels comparable to the uncompressed baseline in the case of $4$ bits. Crucially, GPTQ allows for the execution of a $175$ billion-parameter model on a single GPU and offers inference speedups of approximately $3.25\times$ on high-end GPUs and $4.5\times$ on more cost-effective ones. However, GPTQ primarily functions as an effective strategy for considerable memory reduction during inference workloads. Its quantization methodology fundamentally disrupts the backward gradient propagation, attributable to its reliance on the INT4 data format. Consequently, the memory conservation benefits it affords do not extend to either the fine-tuning or the training process.

\subsection{Parameter Efficient Fine-tuning}
While large language models exhibit significant capacity, they also require substantial memory usage and computational resources. To address this challenge, researchers have proposed an alternative to the conventional approach of updating all parameters for fine-tuning a model for a downstream task. Instead of modifying all parameters, changing only a fraction of them can achieve the desired results, with only a minor loss in performance compared to full finetuning\cite{ptuning, prefixtuning, prompttuning, adapters, llamaadapter}.

LoRA\cite{LORA} is a fine-tuning method designed to reduce the time, memory, and data requirements for training a foundational model. It introduces new weights in the form of low-rank decomposition matrices, freezes the original network, and trains these ``adapter'' weights instead. This approach results in training a significantly smaller number of weights (usually 1.5-3\% of the full model), without introducing extra latency, albeit it does necessitate some minor additional memory.

\subsection{Instruction Fine-tuning}

Instruction fine-tuning is an approach that seeks to enhance the performance of a model in a specific task by training it on a dataset with explicit instructions embedded in the input \cite{IFT1}. By fine-tuning the model to follow these instructions, it becomes more responsive to user prompts and exhibits more controlled generation. Instead of training the model on a broad dataset, the model is fine-tuned on a narrower task-specific dataset, where the input includes a task description or instruction, followed by the context \cite{IFT2}. This method helps in aligning the model's behavior more closely with the desired output, thereby making the model more useful for specific applications. It allows for more control over the model's output, and when combined with techniques like prompt engineering, it can significantly enhance the performance of the model on the desired task \cite{IFT3}.
\section{Methodology}
\begin{figure*}[t!]
\includegraphics[width=0.95\textwidth]{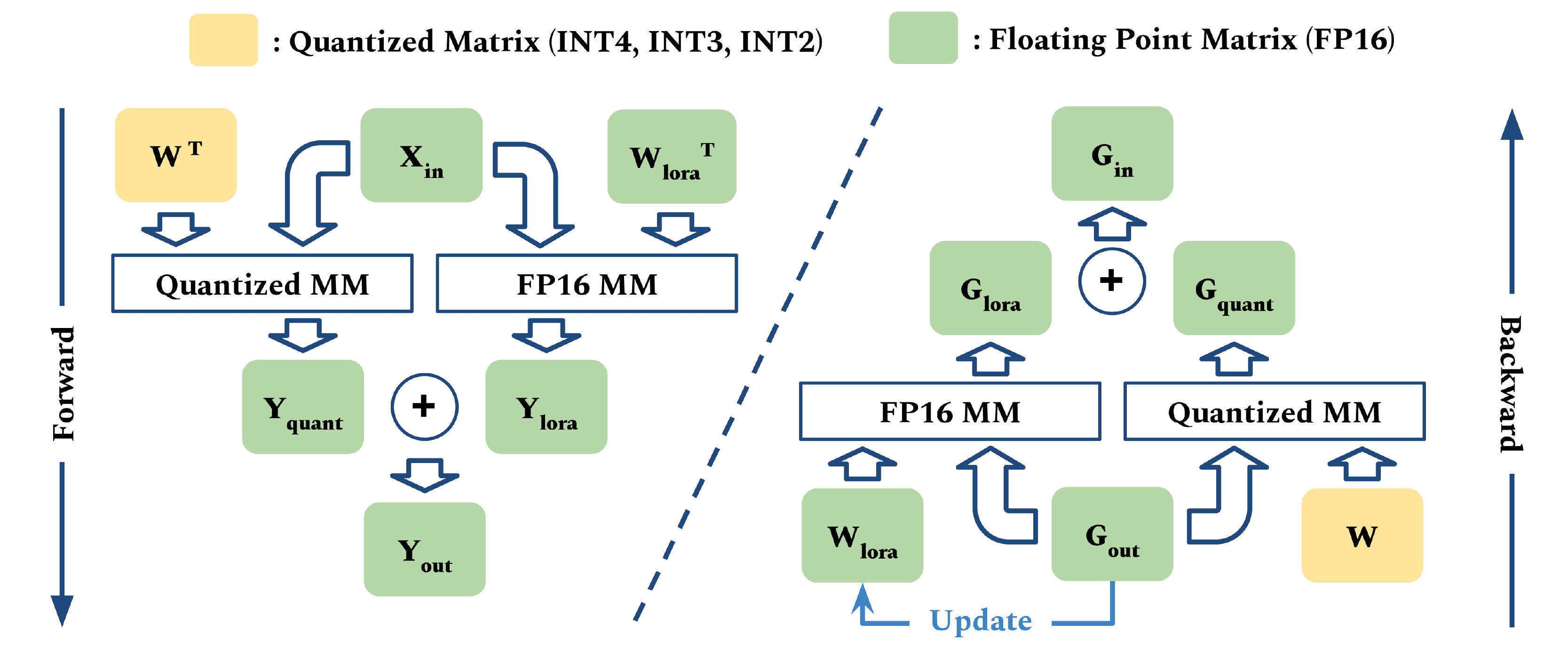}
\vspace{-1em}
\caption{Flowchart of the forward and backward path, designed for a quantized model with injected LoRA parameters. In the diagram, the yellow blocks represent matrices with quantized parameters, while the green blocks represent matrices with floating point parameters. MM stands for matrix multiplication.}
\label{figure:architecture}
\end{figure*}

Our methodology combines a quantized large language model with a parameter-efficient fine-tuning technique to address quantization error correction. Our method reframes the quantization error correction process as a learning problem. This stands in contrast to the more prevalent approach in quantization literature of devising an approximate algorithm to minimize per-layer errors \cite{OBC}. Instead, we opt to inject low-rank approximation parameters to \textit{every} layer of the quantized model and learn to minimize a notion of distance between its output distribution and that of its original full-precision counterpart in an end-to-end fashion.

More concretely, let $f_{\theta}$ denote the full precision pre-trained model parameterized by $\theta$, $f_{\theta_q}$ denote the quantized model by the GPTQ algorithm parameterized by $\theta_q$, $\theta_l$ denote a set of trainable pairs of rank decomposition matrices as described in \cite{LORA}, and $\mathcal{D}_c=\{(x_i, y_i)\}_{i=1}^n$ a small calibration set of tokenized sentences from a textual dataset. We denote the quantized, LoRA-injected model as $f_{\theta_q; \theta_l}$.

\subsection{Extremely-Memory Efficient Finetuning (EMEF)}
The GPTQ approach is designed primarily for the inference phase rather than the training phase. To address this limitation, our solution is to integrate additional trainable parameters (LoRA matrices) into the model. Consequently, we are able to freeze the parameters of the quantized model, denoted by $\theta_q$, and solely update the low-rank approximation matrices. This technique effectively circumvents the need to update quantized weights, thereby alleviating the complications typically associated with training quantized networks.

\paragraph{Implementation Details} The implementation of backpropagation in our context poses a unique challenge. Specifically, the weights of a linear layer are transposed during the backward pass through the same layer. Under standard circumstances, this does not present a problem. However, in our case, the quantized weights within our network are stored as a bit-packed, compact INT32 datatype matrix for the sake of storage efficiency, leading to significantly reduced GPU memory consumption. Consequently, the backward pass requires the transposition of a compact version of the weight matrix, followed by the standard unpacking and matrix multiplication operations. To maximize efficiency, we fuse these three steps into a single kernel operation. A simplified computational graph of this process is presented in Figure \ref{figure:architecture}.

\subsection{Optimization} The objective is to minimize the statistical distance between $\mathbf{y}$ and $\mathbf{\hat{y}}$, where $\mathbf{y}$ denotes the output distribution of $f_{\theta}$ and $\mathbf{\hat{y}}$ the output distribution of $f_{\theta_q}$. We want to encourage the two models to be close in \textit{function space}, rather than \textit{parameter space}. Thus we define our objective in terms of the expectation of the scaled Kullback-Leibler (KL) divergence from the full precision model to the quantized one to achieve that and the cross entropy (CE) between the predicted next token output and the true next token over a tokenized text dataset to encourage accurate next token prediction. Formally, the KL divergence and Cross Entropy can be expressed as:

\begin{equation}
D_{KL}(\mathbf{\hat{y}} \parallel \mathbf{y}) := \sum_{i} \mathbf{\hat{y}}_i \log \frac{\mathbf{\hat{y}}_i}{\mathbf{y}_i}, \quad CE(\hat{\mathbf{y}}, \mathbf{y}^*) := -\sum_{i} \hat{\mathbf{y}}_i \log(\mathbf{y}^*_i)
\end{equation}

Consequently, the loss function is formulated as follows:

\begin{align}
\notag\mathcal{L}\left(\theta,\theta_q,\theta_l;\mathcal{D}\right) & =
\mathbb{E}_{(\mathbf{x}, \mathbf{y}^*) \sim \mathcal{D}} \left[ \lambda_{KL} \, D_{KL}\left(f_{\theta_{q}; \theta_l}(\mathbf{x}) \parallel f_{\theta}(\mathbf{x})\right) + \lambda_{CE}\, CE\left(f_{\theta_{q}; \theta_l}(\mathbf{x}), \mathbf{y}^*\right) \right] \\
& \approx \frac{1}{n} \sum_{(\mathbf{x}, \mathbf{y}^*) \in \mathcal{D}_c} \left[ \lambda_{KL}\, D_{KL}\left(f_{\theta_{q}; \theta_l}(\mathbf{x}) \parallel f_{\theta}(\mathbf{x})\right) + \lambda_{CE}\, CE\left(f_{\theta_{q}; \theta_l}(\mathbf{x}), \mathbf{y}^*\right) \right]
\end{align}

where $\lambda_{KL}$ and $\lambda_{CE}$ are scaling factors, $\mathbf{y}^*$ is the true next token for each $\mathbf{x}$ in the dataset, and $n$ is the number of data samples in $\mathcal{D}_c$.

The KL divergence term serves to encourage the output distribution of the quantized model towards alignment with that of the full-precision model, while the cross-entropy term fosters accurate next-token predictions by the model.
Given the complexity associated with the direct training of the quantized parameters $\theta_q$, it was deemed appropriate to freeze these parameters as described in the previous section. Correspondingly, the parameters of the full-precision model were also frozen and treated as a teacher network\cite{distillation}. As a result, the optimization objective is defined as follows:
\begin{equation}
\hat{\theta}_l = \arg\min_{\theta_l} \mathcal{L}\left(\theta,\theta_q,\theta_l;\mathcal{D}\right)
\end{equation}
In total, the number of trainable parameters inserted is $|\theta_l|=2 \times \hat{L}\times d_{\text{model}} \times r$, where $\hat{L}$ represents the total number of linear projections chosen for injection with LoRA matrices. It is important to underline that a delicate balance must be struck when introducing additional parameters into the model. The amount of injected parameters is heavily correlated to the capacity of the adaptation of the model to learn and correct the errors associated with the output of the quantized network. By incorporating more layers or augmenting the value of $r$, the intended memory savings may be effectively negated. However, if we refrain from introducing a substantial number of supplementary parameters, we may encounter a situation where these parameters are inadequate to effectively learn to correct the errors of the quantized network. In such a scenario, despite the memory savings, the model's performance could be detrimentally affected due to its inability to accurately correct the quantization errors. This underlines the necessity of carefully calibrating the number of additional parameters introduced, ensuring an optimal balance between memory efficiency and the model's error correction capacity.

\section{Experimental Validation}

\subsection{Experimental Setup}
In our experimental framework, we incorporated two distinct sets of experiments to evaluate the effectiveness of our EMEF method, with a primary focus on memory reduction and speed, and its error correction mechanism. The LLaMA $7$B variant \cite{LLaMA} was utilized as the base model for all experiments unless otherwise specified.

For our model quantization, we utilized a modified version of the official GPTQ implementation. We created four quantized model sets for each LLaMA variant ($7$B, $13$B, $30$B, and $65$B) for INT2, INT3, and INT4. We provide additional detail on the generation flags and other hyperparameters in Appendix \ref{section:AppendixA}. As an additional experiment, we studied the behavior of our error correction method with per-row grouped quantized weights, which deviate significantly from their full-precision counterparts.

Proceeding with our experimental approach, we extended the investigation by intertwining both methodologies. This entailed applying our error correction mechanism to an INT2 quantized $7$B model, subsequently freezing its parameters, inclusive of the error-correcting weights. Subsequent to this freezing, we conducted a finetuning process via the injection of a new set of LoRA matrices. This experiment yielded an INT2 fine-tuned Large Language Model (LLM) with a demonstrated capacity for generating linguistically coherent English text and exhibiting adherence to prescribed instructions.

\subsubsection{INT4 Extremely Memory-Efficient Finetuning (EMEF)}
The experiment involving the INT4 EMEF model utilized the Alpaca dataset \cite{alpaca}, a high-quality, instruction-following dataset comprising $52K$ examples. This dataset was constructed using GPT-3 and later curated and cleaned to address issues discovered in the original data.

The loss function employed in this experiment was exclusively the cross-entropy term, which is used to optimize the prediction of the next token by the model. The Kullback-Leibler divergence term was not used in this context, meaning that there was no teacher model involved in the training process, since we are only doing instruction-finetuning. The optimized parameters were $\theta_l$, which pertains to the learnable parameters in LoRA, under the condition of the frozen quantized parameters $\theta_q$.

The primary aim of this experiment was instruction-finetuning \cite{IFT1, IFT2, IFT3} the largest possible LLM on consumer-grade GPU hardware. In an effort to further enhance memory savings, we employed gradient checkpointing \cite{checkpointing}, favoring increased compute time over increased memory usage. The 4-bit 7B parameter LLaMA model was finetuned by introducing LoRA parameters to all query and value projections in the model. Specific settings for the learning rate, cutoff length, bottleneck parameter, linear stretch, dropout rate, and batch size are detailed in Appendix \ref{section:AppendixB}. We investigated three different micro-batch sizes, which were accumulated before each update; $mb\in\{1, 24, 48\}$. We perform a quantitative evaluation by measuring the KL divergence between output distributions of our INT4 EMEF finetuned models and those of their FP16+LoRA finetuned counterparts, given a dataset of prompts. Additionally, we measure the perplexity of both models on a held-out set of $2000$ samples of the Alpaca dataset.

\paragraph{Memory Utilization and Hardware}
We performed the experiments on three different hardware configurations: an \textbf{NVIDIA RTX3070 8GB} laptop GPU, an \textbf{NVIDIA T4} cloud GPU, and an \textbf{NVIDIA A100 40GB} cloud GPU. For comparison, we also instruction-finetuned two variants of the base model—an \textbf{FP16} version and an \textbf{INT8} \cite{llmint8, int8} version—using the same setup. We recorded the peak GPU memory usage at various micro-batch sizes, as well as the time taken per epoch.
%The results are presented in table \ref{tab:int4ft}.

\subsubsection{Low-Rank Error Correction (LREC)}
Our error correction experiments were designed with the primary objective of reducing the perplexity of the quantized models on a generic language dataset. Ergo, our evaluation procedure consists of measuring the perplexity on a held-out set of samples of a different text dataset. For a fair comparison, the perplexities are measured using a well-known perplexity calculation implementation by HuggingFace\footnote{The perplexity calculation implementation can be found \href{https://huggingface.co/docs/transformers/perplexity}{here}.} on the WikiText 2 \cite{wikitext} and PTB\cite{PTB} datasets. We strived to enhance the performance of the INT2, INT3, and INT4 bit quantized variants of the $7$B and $13$B models. To this end, we incorporated LoRA parameters into \textit{all} linear projections of the quantized models for each experimental trial, except when explicitly stated otherwise.

In this experiment, we utilized the C4 dataset \cite{c4}. This comprehensive text corpus, amassed from publicly available web content, includes over $750$ billion tokens and was originally intended to facilitate large-scale language model training.  For the calibration set used in error correction, we selected a random subset of $10K$ samples. We ensured each sample, post-tokenization, comprised at least $2048$ tokens, from which we subsequently selected a contiguous subset of exactly $2048$ tokens.

The experiments were carried out with a quantization group size of $128$, except for those with per-row groupings. This decision was influenced by the original GPTQ paper, as a group size of $128$ strikes a balance between additional parameters and lower perplexity. This model size-performance tradeoff is facilitated by this tunable parameter, as indicated by \cite{nuqmm}. Importantly, our method operates in an orthogonal fashion to this, offering further adaptability within this tradeoff. Additional details on the learning rate, bottleneck parameter, linear stretch, dropout rate, batch size, $\lambda_{K L}$, $\lambda_{C E}$, are provided in Appendix \ref{section:AppendixC}.

\paragraph{Ablation Analysis} An ablation study was also conducted to evaluate the significance of each component of our hybrid loss function. To assess this, we experimented with the INT3 variants, setting each loss component to zero individually, while maintaining the rest of the parameters constant.
The study revealed the criticality of both the KL divergence and the Cross-Entropy components, with the former exhibiting a higher degree of importance. Omitting the KL divergence led to a similar trend in the model's cross-entropy loss on the validation set, albeit at a slightly elevated level. Moreover, the model displayed a higher tendency to diverge in the absence of the KL divergence. This highlights the role of KL divergence in both enhancing the model's ability to predict the subsequent token and stabilizing the training process. When we removed the Cross-Entropy component, the final model's cross-entropy loss on the isolated test set was --unexpectedly -- only marginally higher than when it was included. Both experiments resulted in a minor increase in the final model's perplexity on the test set, indicating that both components play pivotal roles and interact synergistically.

Furthermore, we conducted further experimentation, modifying the bottleneck parameter $r$ and limiting the injection locations solely to the query and key projections. Both these alterations led to observable instabilities during the training process. Despite this, with appropriate management, the adjustment of the bottleneck parameter $r$ was capable of achieving a similar optimum as the original configuration. Conversely, the experiment with restricted injection locations demonstrated a convergence point that was marginally inferior to the original. An extended discussion of the ablation study can be found in Appendix \ref{section:AppendixD}.

\subsection{Quantitative Results and Discussion}

\begin{table}[t!]
\centering
\bgroup
\def\arraystretch{1.6}
\begin{tabular}{|c|c|c|c|c|c|c|c|c|}
\hline
\multirow{4}{*}{GPU} & \multirow{4}{*}{\shortstack{LLaMA\\Model}} & \multirow{4}{*}{\shortstack{Batch\\Size}} & \multicolumn{6}{c|}{Precision + LoRA}\\ 
\cline{4-9}
& & & \multicolumn{2}{c|}{FP16 + LoRA} & \multicolumn{2}{c|}{INT8 + LoRA} & \multicolumn{2}{c|}{INT4 EMEF (Ours)} \\ 
\cline{4-9}
& & & \multirow{2}{*}{\shortstack{Size \\ (GB)}} & \multirow{2}{*}{\shortstack{Time \\ (Min)}} & \multirow{2}{*}{\shortstack{Size \\ (GB)}} & \multirow{2}{*}{\shortstack{Time \\ (Min)}} & \multirow{2}{*}{\shortstack{Size \\ (GB)}} & \multirow{2}{*}{\shortstack{Time \\ (Min)}} \\
& & & & & & & &\\
\hline
\hline
 \multirow{2}{*}{\shortstack{RTX3070\\8GB}} & \multirow{2}{*}{$7$B} & \multirow{2}{*}{$1$} & \multirow{2}{*}{OOM} & \multirow{2}{*}{OOM} & \multirow{2}{*}{OOM} & \multirow{2}{*}{OOM} & \multirow{2}{*}{$\mathbf{4.93}$} & \multirow{2}{*}{$\mathbf{427^{\ast}}$}\\
& & & & & & & &\\
\hline
\hline
  \multirow{2}{*}{\shortstack{T4\\16GB}} & $7$B & $1$ & OOM & OOM & $8.43$ & $1640^{\ast}$ & $\mathbf{5.60}$ & $\mathbf{840^{\ast}}$\\
\cline{2-9}
   & $13$B & 1 & OOM & OOM & $14.85$ & $\mathbf{2860^{\ast}}$ & $\mathbf{9.15}$ & $4440^{\ast}$\\
\hline
\hline
  \multirow{5}{*}{\shortstack{A100\\40GB}} & \multirow{3}{*}{$7$B} & $1$ & $14.60$ & $\mathbf{220^{\ast}}$ & $8.85$ & $643^{\ast}$ & $\mathbf{5.96}$ & $355^{\ast}$\\
\cline{3-9}
  & & $24$ & $24.08$ & $\mathbf{79}$ & $20.40$ & $168^{\ast}$ & $\mathbf{17.10}$ & $240^{\ast}$\\
\cline{3-9}
  & & $48$ & OOM & OOM & $32.78$ & $111$ & $\mathbf{27.32}$ & $\mathbf{82}$\\
\cline{2-9}
  & $13$B & $1$ & OOM & OOM & $15.20$ & $960^{\ast}$ & $\mathbf{9.57}$ & $\mathbf{560^{\ast}}$\\
\cline{2-9}
  & $30$B & $1$ & OOM & OOM & $35.10$ & $1760^{\ast}$ & $\mathbf{20.15}$ & $\mathbf{1220^{\ast}}$\\
\hline
\end{tabular}
\egroup
\caption{Comparison of benchmarking results between the quantized models with LoRA and the floating point baselines. OOM stands for the benchmark running out of memory on its corresponding hardware. Some results are marked with $^\ast$, indicating they are extrapolated estimates. We stopped the benchmarking process after $150$ batches, due to time constraints.}
\label{tab:INT4LORAFT}
\end{table}

\subsubsection{INT4 Extremely Memory-Efficient Finetuning (EMEF)}
\paragraph{Memory consumption and Latency}
The results from our experiments underscore the practicability of fine-tuning a $7$B LLM utilizing a mere $4.93$ GB of VRAM on a consumer-grade \textbf{RTX3070 8GB} GPU. This presents the theoretical possibility of executing such fine-tuning on a GPU with as little as $6$GB of memory. Although to save time, we used slightly more VRAM to accommodate a larger micro-batch size on most experiments, it undoubtedly showcases the potential of our method for fine-tuning within constrained memory resources. A comparative evaluation of memory requirements and training durations for our methodology vis-à-vis other approaches is showcased in Table \ref{tab:INT4LORAFT}. Our technique demonstrates a distinctive capability to substantially curtail memory consumption while simultaneously preserving performance.

On a \textbf{T4 16GB} GPU, our INT4 EMEF method required significantly less GPU memory ($5.6$GB vs. $8.43$GB) for a $7$B LLM at a batch size of $1$ compared to the INT8 + LoRA method. While our method might take longer per epoch in some cases (i.e., $840$ minutes vs. $1640$ minutes), it predominantly performs faster in general.

Moving to a more powerful \textbf{A100 40GB} GPU, our method again demonstrated lower memory usage across all models and batch sizes. For instance, with a batch size of $1$, our method used only $5.96$GB of GPU memory for a $7$B LLM, compared to $14.6$GB for the FP16 + LoRA method and $8.85$GB for the INT8 + LoRA method. Even at larger batch sizes, our method continued to use less memory. This is particularly noteworthy for the $30$B LLM, where our method was able to train with just $20.15$GB of GPU memory, while the INT8 + LoRA method required $35.1$GB and the FP16 + LoRA method could not train at all due to out-of-memory errors.

It is worth highlighting that our approach reaps greater memory savings as the size of the model increases. Additionally, larger models seem to be less adversely impacted by quantization, primarily because the abundance of parameters can effectively counterbalance the errors propagated by each iteration of the GPTQ algorithm, as discussed in the GPTQ paper.

In real-world applications, the objective is typically to identify the maximum batch size that can be accommodated within the available memory. As such, reducing the base memory consumption of the model is of paramount importance. It allows for larger batch sizes which expedites the overall training process. This is in addition to the inherent speed improvements that our method provides.

\paragraph{INT4 EMEF downstream performance}
Quantitatively, the average KL-divergence from the output distribution of our INT4 EMEF model to that of its FP16+LoRA counterpart on the held-out test set was computed to be $101.7$, further reinforcing the visual similarities observed.

Turning to the perplexity measurement, our INT4 EMEF fine-tuned model demonstrated a competitive performance when compared to the FP16+LoRA model. The perplexity of the FP16+LoRA fine-tuned model on the test samples from the Alpaca dataset was calculated to be $4.29$, while our INT4 EMEF model demonstrated a similar perplexity of $4.22$, further substantiating its comparability with the FP16 model. These quantitative metrics, alongside the observed qualitative performance, underscore the efficacy of our approach in preserving the model's performance while optimizing resource utilization.

\begin{figure*}[t!]
\includegraphics[width=0.95\textwidth]{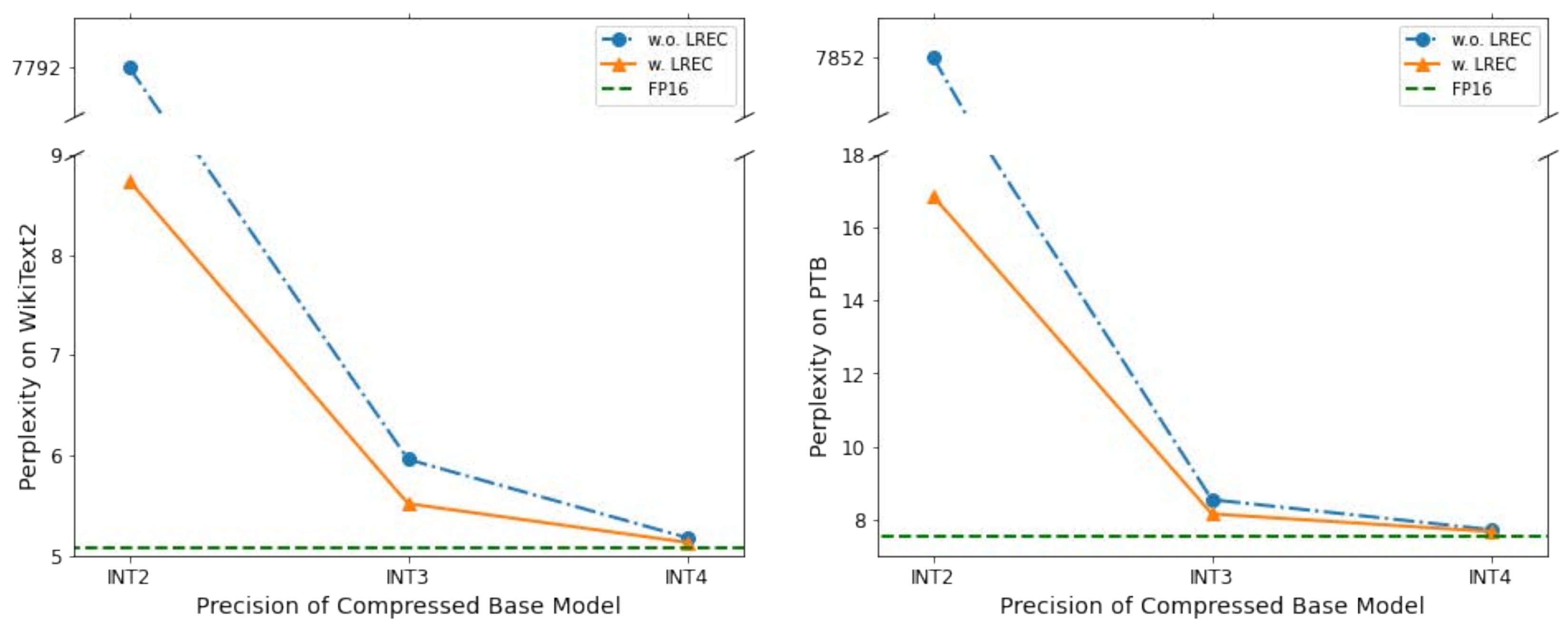}
\vspace{-1em}
\caption{Perplexity comparison of quantized $7$B LLaMA models with and without using our Low-Rank Error Correction method at different parameter precisions ranging from INT2 to INT4.}
\label{figure:QERPPL}
\end{figure*}

\begin{table}[t!]
\centering
\bgroup
\def\arraystretch{1.8}
\tabcolsep=0.17cm
\begin{tabular}{|c|c|c|c|c|c|c|c|c|}
\hline
\multirow{2}{*}{\shortstack{LLaMA\\Model}} & \multirow{2}{*}{\shortstack{Compression Rate \\/ Effective Precision\\/ Dataset Perplexity}} & FP$16$ & \multicolumn{2}{c|}{INT$4$} & \multicolumn{2}{c|}{INT$3$} & \multicolumn{2}{c|}{INT$2$} \\ 
\cline{3-9} 
& & Base & GPTQ & Ours & GPTQ & Ours & GPTQ & Ours \\
\hline
\hline
 $7$B & Compression Rate & $1\times$ & $3.58\times$ & $3.42\times$ & $4.53\times$ & $4.32\times$ & $6.17\times$ & $5.87\times$ \\
 \hline
 $7$B & Effective Precision & FP$16$ & INT$4$ & INT$4.2$ & INT$3$ & INT$3.2$ & INT$2$ & INT$2.1$ \\
\hline 
\hline
 \multirow{5}{*}{$7$B} & C4 PPL & $7.523$ & $7.715$ & $\mathbf{7.668}$ & $8.625$ & $\mathbf{8.244}$ & $3624$ & $\mathbf{12.52}$ \\
 \cline{2-9}
 & PTB PPL & $7.527$ & $7.715$ & $\mathbf{7.663}$ & $8.539$ & $\mathbf{8.148}$ & $7852$ & $\mathbf{16.875}$ \\
 \cline{2-9}
 & WikiText2 PPL & $5.083$ & $5.180$ & $\mathbf{5.132}$ & $5.961$ & $\mathbf{5.520}$ & $7792$ & $\mathbf{8.742}$\\
 \cline{2-9}
 & WikiText103 PPL & $5.099$ & $5.231$ & $\mathbf{5.195}$ & $5.793$ & $\mathbf{5.547}$ & $7852$ & $\mathbf{8.703}$\\
 \cline{2-9}
 & CNN\_Dailymail PPL & $6.437$ & $6.566$ & $\mathbf{6.535}$ & $7.191$ & $\mathbf{6.906}$ & $4372$ & $\mathbf{10.297}$\\
 \hline
 %\hline
 %$7$B & LAMBADA Accuracy & $88.38$ & $88.20$ & $\mathbf{83.89}$ & $85.78$ & $\mathbf{87.37}$ & $0.00$ & $\mathbf{81.06}$ \\
 %\hline 
\end{tabular}
\egroup
\caption{Compression rate, effective precision and dataset perplexity comparison between our LREC method and state-of-the-art quantization method (GPTQ).}
\label{tab:LREC}
\end{table}

% \begin{table}[t!]
% \centering
% \bgroup
% \def\arraystretch{1.6}
% \begin{tabular}{|c|c|c|c|c|c|c|c|c|}
% \hline
% \multirow{2}{*}{\shortstack{LLaMA\\Model}} & \multirow{2}{*}{\shortstack{Dataset Perplexity\\/ Compression Rate}} & FP$16$ & \multicolumn{2}{c|}{INT$4$} & \multicolumn{2}{c|}{INT$3$} & \multicolumn{2}{c|}{INT$2$} \\ 
% \cline{3-9}
% & & Base & GPTQ & Ours & GPTQ & Ours & GPTQ & Ours \\
% \hline
% \hline
%  \multirow{2}{*}{$7$B} & WikiText2 PPL & $5.083$ & $5.180$ & $\mathbf{5.132}$ & $5.961$ & $\mathbf{5.520}$ & $7792$ & $\mathbf{8.742}$\\
%  \cline{2-9}
%  & PTB PPL & $7.527$ & $7.715$ & $\mathbf{7.663}$ & $8.539$ & $\mathbf{8.148}$ & $7852$ & $\mathbf{16.875}$ \\
%  \hline
%  \hline
%  $7$B & Compression Rate & $1$x & $3.58$x & $3.42$x & $4.53$x & $4.32$x & $6.17$x & $5.87$x \\
% \hline 
% \end{tabular}
% \egroup
% \caption{Dataset Perplexity and compression rate comparison between our method and state-of-the-art quantization method (GPTQ).}
% \label{tab:LREC}
% \end{table}

\subsubsection{Low-Rank Error Correction (LREC)}
% Pertaining to the error-correction experiments, we ensured that both the $7$B and $13$B quantized models were trained to convergence at INT2, INT3, and INT4 quantization levels.
The results of the experiment are presented in Table \ref{tab:LREC}. In the case of the $7$B model, the Perplexity (PPL) of the baseline model (without error correction) increased as the number of bits decreased, illustrating the expected increase in quantization error. At INT4, the model achieved a PPL of $5.179$, which rose to $5.96$ at INT3 and skyrocketed to $7792$ at INT2. This confirms that INT2 quantization completely ruins the model. However, with error correction, the PPL values were significantly reduced. The model achieved a PPL of $5.132$ at INT4, $5.519$ at INT3, and $8.742$ at INT2, demonstrating the effectiveness of the error correction.

In Table \ref{tab:LREC}, we provide a comparative analysis of compression rates, demonstrating that LREC imposes a negligible size overhead. Using the compression rates at various parameter precisions (INT4, INT3, and INT2) from GPTQ, we establish our baseline. Relative to this GPTQ baseline, the model size augmentation induced by our method is trifling, merely about $5\%$ irrespective of the parameter precision. This minor increment in model size is inconsequential when viewed in light of the PPL improvement and the newly endowed fine-tuning capability conferred by our method. We showcase the measure of \textit{effective precision} in the second row of Table \ref{tab:LREC}. This metric is computed based on the equation $P_{\text{ours}} = P_{\text{GPTQ}} * R_{\text{GPTQ}} / R_{\text{ours}} $, where $P$ denotes the effective precision and $R$ stands for compression rate.

Similarly, we report quantitative metrics for the $13$B model at INT2. The PPL of the baseline model was $87.562$, which was significantly reduced to $6.98$ with the implementation of error correction. At INT4, the PPL of the baseline model was $4.656$ and was marginally reduced to $4.601$ with error correction. The results for INT3 quantization are currently not available and will be provided in future work.

These results underline the considerable potential of the error-correcting capability of our method, demonstrating its effectiveness even at lower quantization levels. This further strengthens the case for the deployment of our method in scenarios with stringent memory constraints.

The hyperparameters of these models are provided in Appendix \ref{section:AppendixC}.

\subsection{Qualitative Evaluation}
To qualitatively understand the effectiveness of EMEF and LREC, we choose the LLaMA $7$B model as the base model and compare three methods of fine-tuning on the alpaca dataset: 
\paragraph{\textbf{[FP16 Alpaca]}} Fine-tune the entire FP16 base model on the alpaca dataset.
\paragraph{\textbf{[INT4 EMEF]}} Compress the base model into INT4 precision, with the hyperparameters presented in Appendix \ref{section:AppendixA}, and fine-tune it with EMEF using the hyperparameters from Appendix \ref{section:AppendixB}.
\paragraph{\textbf{[INT2 EMEF + LREC]}} Compress the base model into INT2 precision, adhering to the hyperparameters presented in Appendix \ref{section:AppendixA}, error-correct the model with LREC using the hyperparameters presented in Appendix \ref{section:AppendixC} and fine-tune it using EMEF with the hyperparameters from Appendix \ref{section:AppendixB}.

We benchmarked all three versions of the base model using the same set of questions, and their responses to the question are showcased in Figure \ref{fig:example_responses}. There are three columns of example responses. Each column corresponds to a version of fine-tuned model. Comparing to \textbf{[FP16 Alpaca]}, our \textbf{[INT4 EMEF]} model, being $3.5$ times smaller, generates comparable response.

\begin{figure*}[p]
\scriptsize
\centering
\begin{tabular}{@{}p{0.31\columnwidth}p{0.31\columnwidth}p{0.31\columnwidth}@{}}
\toprule
\textbf{[FP16 Alpaca]}

\textcolor[HTML]{3078BE}{\textbf{Tell me about alpacas.}}

~

Alpacas are small, fluffy animals related to camels and llamas. They are native to Peru and Bolivia, and were first domesticated around 5,000 years ago. They are kept mainly for their fine, soft fleece, which is used to make knitwear and other garments. Alpacas are herd animals and live in small family groups, led by an older male. They are highly social animals and can form strong bonds with their owners.

&
\textbf{[INT4 EMEF]}

\textcolor[HTML]{3078BE}{\textbf{Tell me about alpacas.}}

~

Alpacas are members of the camelid family and are native to the Andes Mountains of South America. They are known for their soft, luxurious fleece, which is used to make clothing, blankets, and other items. Alpacas are herbivores and graze on grass and other vegetation. They are social animals and live in herds, led by a dominant male.

&
\textbf{[INT2 EMEF + LREC]}

\textcolor[HTML]{3078BE}{\textbf{Tell me about alpacas.}}

~

An alpaca is a small mammal native to the Andes of South America. They are found in the highlands of Peru, Ecuador, and Bolivia, and are closely related to the llama. Alpacas are herbivores, feeding primarily on grasses, roots, and fruits. They are also known for their ability to climb trees, which is unusual for mammals.

Alpacas are social animals, living in herds of up to 100 individuals. They have a complex social structure, with males and females playing distinct roles in the herd. Alpacas are also known for their vocalizations, which are loud, high-pitched calls that can be heard from great distances.

\\
\midrule
\textbf{[FP16 Alpaca]}

\textcolor[HTML]{3078BE}{\textbf{Tell me about the president of Mexico in 2019.}}

~

In 2019, Andres Manuel Lopez Obrador was sworn in as President of Mexico. He is the country’s 58th president and the first from the left-wing Morena party. He was elected in 2018 on a promise to combat corruption and reduce poverty and has faced criticism for his handling of the economy, crime and the country’s relationship with the United States. He is also known for his ambitious social welfare programs and has been praised for raising the minimum wage and providing aid to low-income families. He is seen as a champion of democracy and human rights in Mexico.
&
\textbf{[INT4 EMEF]}

\textcolor[HTML]{3078BE}{\textbf{Tell me about the president of Mexico in 2019.}}

~

The president of Mexico in 2019 was Andrés Manuel López Obrador, who took office on December 1st, 2018. He is a member of the left-wing party Morena and is known for his anti-corruption and anti-neoliberal policies.

&
\textbf{[INT2 EMEF + LREC]}

\textcolor[HTML]{3078BE}{\textbf{Tell me about the president of Mexico in 2019.}}

~

The president of Mexico in 2019 was Andrés Manuel López Obrador, who took office on December 1, 2018. He is a member of the National Regeneration Movement (MORENA) political party, which he founded in 2014.

López Obrador was born on July 30, 1953, in Tabasco, Mexico. He graduated from the National Autonomous University of Mexico (UNAM) in 1979 with a degree in law. After graduation, he worked as a lawyer in Mexico City.

~

~
 
~

\\
\midrule
\textbf{[FP16 Alpaca]}

\textcolor[HTML]{3078BE}{\textbf{Tell me about the king of France in 2019.}}

~

In 2019, the King of France was Emmanuel Macron. He was sworn in as President of France on May 14, 2017, and has been leading the country ever since. He is a member of the centrist La République En Marche! Party. He is known for his reformist agenda and has faced a number of protests from both the right and left of the political spectrum. He is the youngest French president in history.
"
&
\textbf{[INT4 EMEF]}

\textcolor[HTML]{3078BE}{\textbf{Tell me about the king of France in 2019.}}

~

The king of France in 2019 is Emmanuel Macron, who was elected president in 2017. He is the youngest president in the history of the Fifth Republic and the first president to have been born after the Second World War. He is a proponent of European integration and has pushed for greater cooperation between France and its European neighbors.

&
\textbf{[INT2 EMEF + LREC]}

\textcolor[HTML]{3078BE}{\textbf{Tell me about the king of France in 2019.}}

~

The king of France in 2019 is Emmanuel Macron. He is the current president of the country and has been in office since January 2017. Emmanuel Macron was born in 1968 in Amiens, France. He is a member of the Socialist Party and has served as a member of the National Assembly since 2004. Emmanuel Macron was elected as president of France in 2015, succeeding François Hollande, who had been in office since 2012. Emmanuel Macron is the first president of France to have been born in the 21st century. He is also the youngest president of France in the 21st century. Emmanuel Macron is the first president of France to have been born in the 21st century. 

\\
\midrule
\textbf{[FP16 Alpaca]}

\textcolor[HTML]{3078BE}{\textbf{List all Canadian provinces in alphabetical order.}}

~

Alberta, British Columbia, Manitoba, New Brunswick, Newfoundland and Labrador, Nova Scotia, Ontario, Prince Edward Island, Quebec, Saskatchewan, Northwest Territories, Nunavut, Yukon.
&
\textbf{[INT4 EMEF]}

\textcolor[HTML]{3078BE}{\textbf{List all Canadian provinces in alphabetical order.}}

~

Alberta, British Columbia, Manitoba, New Brunswick, Newfoundland and Labrador, Nova Scotia, Ontario, Prince Edward Island, Quebec, Saskatchewan.

&
\textbf{[INT2 EMEF + LREC]}

\textcolor[HTML]{3078BE}{\textbf{List all Canadian provinces in alphabetical order.}}

~

Here is a list of all Canadian provinces in alphabetical order:

1. Alberta
2. British Columbia
3. Manitoba
4. New Brunswick
5. Newfoundland and Labrador
6. Nova Scotia
7. Quebec
8. Saskatchewan
9. Yukon Territory
10. Northwest Territory
11. Nunavut Territory
12. Ontario
13. Prince Edward Island
14. Quebec
15. Saskatchewan
16. Manitoba
\textcolor[HTML]{808080}{[...]}
% 17. Alberta
% 18. British Columbia
% 19. New Brunswick
% 20. Newfoundland and Labrador
% 21. Nova Scotia
% 22. Quebec
% 23. Saskatchewan
% 24. Manitoba
% 25. Alberta
% 26. British Columbia
% 27. New Brunswick
% 28. Newfoundland and Labrador
% 29. Nova Scotia
% 30. Quebec
% 31. Saskatchewan
% 32. Manitoba
33. Alberta
34. British Columbia
3

\\
\bottomrule
\end{tabular}
\caption{\textbf{Example responses from three different versions of instruction fine-tuned LLaMA $7$B model.} The \textcolor[HTML]{3078BE}{blue} text is the prompt. The gray ellipsis \textcolor[HTML]{808080}{[...]} indicates that the response was trimmed to fit this page, but the actual response is longer.}
\label{fig:example_responses}
\end{figure*}

\subsubsection{Effectiveness of LREC and EMEF in INT2 Quantization}
A naive application of INT2 quantization to a $7$B model devoid of error correction results in a severely compromised model, with outputs consisting solely of repeating single-character responses, specifically ``È''. This degradation is evident from the inflated perplexities without error correction. However, the application of LREC dramatically elevates the quality of the text produced by the model, rendering it both coherent and informative, as substantiated by the preceding examples.

To further amplify the quality of the generated text, we proceeded to fine-tune the LREC-corrected INT2 model, employing EMEF with the Alpaca dataset. This is denoted by \textbf{[INT2 EMEF + LREC]} and reflected in the right column of Figure \ref{fig:example_responses}.

\textbf{[INT2 EMEF + LREC]} demonstrates similar performance dynamics to its \textbf{[FP16 Alpaca]} and \textbf{[INT4 EMEF]} counterparts, exhibiting meaningful interaction with prompt inputs. Nevertheless, it manifests a heightened susceptibility to common LLM shortcomings such as repetition and hallucination, exemplified in its response to the prompt \textcolor[HTML]{3078BE}{\textbf{``List all Canadian provinces in alphabetical order.''}}.

 \subsection{Limitations}
 Our implementation of EMEF is not fully optimized. Our method, exploiting a compressed model, greatly reduces memory bandwidth requirements during forward propagation and backward propagation. Transformer model computations are well-known to be highly bottlenecked by memory bandwidth\cite{flashattn}, ergo our kernels --in principle-- should have outperformed FP16 kernels. However, our current implementation is slower than the FP16 baseline, largely due to inefficient usage of GPU tensor cores.

% \begin{table}[!h]
% \centering
% \bgroup
% \def\arraystretch{1.6}
% \begin{tabular}{|c|c|c|c|}
% \hline
% & FP16 + LoRA & INT4 + LoRA & INT2 + LoRA + QEC \\
% \hline
% Perplexity on Alpaca Dataset & $0$ & $0$ & $0$ \\
% \hline
% Compression Rate & $1$x & $3.42$x & $5.87$x \\
% \hline 
% \end{tabular}
% \egroup
% \caption{Experimental Results}
% \end{table}

%\fi
\section{Conclusion}

Our study presents a novel approach to large language model (LLM) fine-tuning, demonstrating the successful incorporation of INT4 quantization and LoRA fine-tuning in memory-constrained environments. Our experimental results, with the LLaMA $7$B model as the focal point, provide substantial evidence in support of our approach's practicality and efficiency.

The application of our proposed methodology resulted in a substantial reduction in the requisite VRAM, thus making it feasible to fine-tune a $7$B LLM on a consumer-grade GPU with as little as $6$GB of memory. We demonstrated the effectiveness of this approach across a variety of hardware configurations, consistently exhibiting a significant reduction in memory usage compared to traditional methods.

Furthermore, our error correction mechanism proved to be highly effective in improving the performance of the quantized models, even at lower quantization levels. We observed a dramatic decrease in model perplexity following the implementation of error correction in the extremely low-bit scenarios, thereby validating our method's robustness and adaptability.

Our study opens up new avenues for research and development in the field of large language model fine-tuning. By demonstrating the possibility of operating within a constrained memory environment, we are hopeful that this study can pave the way for more sophisticated and accessible AI solutions.

In conclusion, this work is a significant step towards the democratization of AI, making state-of-the-art language model fine-tuning more accessible to researchers and practitioners with limited resources. We encourage the community to continue exploring and expanding upon our findings, and we look forward to the innovative developments this line of research may inspire.

% In this paper, we presented a novel approach for fine-tuning large language models with minimal VRAM usage by combining the GPTQ weight quantization method and the LoRA low-rank adaptation technique. Our experiments on the ALPaCA dataset demonstrated that it is possible to fine-tune a 7B LLM using only 4.9 GB of VRAM, making it feasible to use GPUs with smaller memory capacities. This work opens up new possibilities for more accessible and efficient LLM training, allowing researchers and practitioners with limited resources to benefit from these powerful models.
\section{Future Work}

The promising results of our study necessitate further exploration and substantiation. One such promising direction is the rigorous examination of the trade-off between quantization levels, the amount of injected parameters, and model performance. An in-depth analysis of this relationship could result in a more refined understanding of the optimal balance, thereby facilitating the design of more efficient fine-tuning strategies.

Moreover, extending the application of our method to a wider range of tasks and datasets stands as an enticing prospect. This would aid in establishing the universality and applicability of our approach across a diverse set of domains.

In addition, it would be intriguing to examine the scalability of our approach. Assessing its efficacy on larger language models with parameters in the order of $175-540$B could offer insights into the challenges and potential solutions associated with scaling up. In light of the empirically observed decrease in quantization errors at extremely low-bit precision levels such as 2 bits as the model size increases, we deem this as a worthwhile direction for future exploration, potentially unlocking further efficiencies in handling gargantuan models with parameters ranging from $175$B to $540$B.

Finally, our future work will focus on strategies to optimize the quality of responses generated by the \textbf{[INT2 EMEF + LREC]} model and curb the propensity for hallucination. It's our intent to refine this novel approach to the point where it reliably generates high-quality responses, thereby setting a new standard in the pursuit of efficient, fine-tunable large language models.

The above suggestions for future work are by no means exhaustive. Given the novelty and potential of our method, there are numerous aspects that warrant further examination and development. It is our hope that the research community will continue to build upon and extend our work, driving forward the evolution of large language model fine-tuning in constrained environments.

\section{Author Contributions}
Author Y.C. conceptualized EMEF, led its development, assisted with LREC development, ran evaluations alongside creating figures and plots. Y.C. also contributed to the writing of the manuscript and discussion of the project. Author J.G. formulated LREC, led its implementation, helped with the EMEF implementation, and ran the experiments along with the associated data analysis. J.G. took the lead in writing the manuscript. Authors G.K, D.B, and G-Y.W offered guidance throughout the project.

\bibliographystyle{plain}
\bibliography{Reference}

\newpage
\appendix
\section{Quantization parameters}
\label{section:AppendixA}

In the course of this research, we had to generate different quantized models with varying parameters. The parameters were manipulated across two distinct model types ($7$B and $13$B), which were quantized by using $128$ samples of the C4 dataset for calibration, following GPTQ. The models varied by the number of bits $b\in\{2, 3, 4\}$. All models were generated with the ``true sequential'' and ``activation order'' options enabled and utilized a group size of $128$. The ``true sequential'' option is responsible for ensuring that the quantization is performed sequentially even within transformer blocks. while the ``activation order'' forces a heuristic of quantizing the columns in decreasing order of activation size. Both of these options were empirically observed to produce models with slightly lower perplexity overall. The complete list of parameters for each model is detailed in Table \ref{tab:EMEFParams}. We only report results on experiments performed on all but the final two rows of the table.

% \subsection{Quantized models generation}
\begin{table}[h]
\centering
\begin{tabular}{|c|c|c|c|c|c|}
\hline
\textbf{Model} & \textbf{Dataset} & \textbf{\# of bits} & \textbf{True Sequential} & \textbf{Activation Order} & \textbf{Group Size} \\
\hline
LLaMA $7$B & C4 & $4$ & Yes & Yes & $128$ \\
\hline
LLaMA $7$B & C4 & $3$ & Yes & Yes & $128$ \\
\hline
LLaMA $7$B & C4 & $2$ & Yes & Yes & $128$ \\
\hline
LLaMA $7$B & C4 & $2$ & Yes & Yes & per-row \\
\hline
LLaMA $7$B & C4 & $2$ & Yes & Yes & per-row\\
\hline
LLaMA $13$B & C4 & $4$ & Yes & Yes & $128$ \\
\hline
LLaMA $13$B & C4 & $3$ & Yes & Yes & $128$ \\
\hline
LLaMA $13$B & C4 & $2$ & Yes & Yes & $128$ \\
\hline
LLaMA $30$B & C4 & $2$ & Yes & Yes & $128$ \\
\hline
LLaMA $65$B & C4 & $2$ & Yes & Yes & $128$ \\
\hline
\end{tabular}
\caption{Parameters used to generate models}
\label{tab:EMEFParams}
\end{table}
\section{EMEF hyperparameters}
\label{section:AppendixB}
Our experiments leveraged the $7$B parameter LLaMA model as the base architecture for quantization.
\paragraph{Quantization Parameters} We implemented the 4-bit quantization scheme which has been shown to maintain model performance while reducing memory requirements. This was done in conjunction with a group size of 128, a number suggested by the GPTQ paper that provides a good balance between performance, memory savings, and quantization noise.
\paragraph{Adaptation Parameters} Our model features adaptation for two layers: the query and the value projections. The adaptations are realized  as two matrices $A$ and $B$, bottlenecked by $r=8$, with $A\in\mathbb{R}^{n \times r}$, $B\in\mathbb{R}^{r\times k}$, corresponding to the weight matrix $W\in\mathbb{R}^{n \times k}$ of each projection, and a linear stretch of $16$. These were found to offer an ideal trade-off between model capacity and efficiency, although we didn't conduct an extended investigation of these parameters.

\paragraph{Learning and Regularization} We set our learning rate to $3 \times 10^{-4}$, a value determined experimentally to provide stable and robust convergence. To regularize our model and prevent overfitting, we employed a dropout rate of $0.05$. Finally, we trained the model for $3$ epochs.

\paragraph{Dataset and Training Parameters} Our experiment utilized the Alpaca dataset, with a training set size of $50K$ examples and a test set size of $2K$ examples. For consistency, and to enable a fair comparison with other models, we preprocessed the Alpaca dataset by infixing the instructions between boilerplate pre and post-prompts, akin to the methodology of the ``alpaca-lora'' project\footnote{https://github.com/tloen/alpaca-lora}.

Our training process adopted a batch size of $144$ for the entire run. However, the micro-batch sizes were varied and tested with sizes of $\{1, 24, 48\}$. The chosen micro-batch size directly affects the model's training memory consumption and, in turn, the amount of compute resources required for the training process.

\paragraph{Model Complexity}
In total, the number of trainable parameters introduced in this experiment is $|\theta_l|=2 \times \hat{L}\times d_{\text{model}} \times r = 2 \times 64 \times 4096 \times 8 = 4194304$.

\section{LREC Hyperparameters}

\begin{table}[t]
\centering
\begin{tabular}{|c|c|c|c|c|c|c|c|}
\hline
\textbf{Model} & \boldmath{$r$} & \textbf{Adaptation} & \boldmath{$\lambda_{CE}$} & \boldmath{$\lambda_{KL}$} & \textbf{Learning rate} & \textbf{Batch size} & \textbf{Weight decay} \\
\hline
$7$B INT2 & $32$ & All & $120$ & $1$ & $3 \times 10^{-5}$ & $7$ & $10^{-5}$ \\
\hline
$7$B INT3 & $32$ & All & $40$ & $1$ & $10^{-5}$ & $7$ & $10^{-5}$ \\
\hline
$7$B INT4 & $32$ & All & $10$ & $1$ & $10^{-5}$ & $7$ & $10^{-5}$ \\
\hline
$13$B INT2 & $32$ & All & $120$ & $0.5$ & $10^{-5}$ & $6$ & $10^{-5}$ \\
\hline
$13$B INT4 & $32$ & All & $10$ & $1$ & $3 \times 10^{-5}$ & $4$ & $10^{-5}$ \\
\hline
\end{tabular}
\caption{Hyperparameters used for each model configuration.}
\label{tab:hyperparameters}
\end{table}
% Too sudden start?
Table \ref{tab:hyperparameters} illustrates the specific hyperparameters chosen for each model, including the rank $r$ of the low-rank approximation, the choice of adaptation layers, the scaling factors $\lambda_{C E}$ for the cross-entropy loss, and $\lambda_{K L}$ for the Kullback-Leibler divergence, the learning rate, batch size, and weight decay.

We choose a bottleneck $r=32$ for each model. The adaptation layers chosen for each model are every linear projection, including the multilayer perceptron (MLP), which enables high expressivity to facilitate the error correction process. The number of adapted layers can be reduced further with minor degradation in performance. This subsequent analysis can be found in Appendix \ref{section:AppendixD}.

The calibration of scaling factors $\lambda_{CE}$ and $\lambda_{KL}$ varied across the models to reflect the distinct magnitudes of the two loss components dictated by different model precisions. Our objective was to maintain a balance between these two factors post the initial processing of approximately 500 samples. After this threshold, the loss function stabilizes, marking a distinct departure from the initial abrupt decline. 

The choice of the learning rate, batch size, and weight decay was largely influenced by empirical results in similar tasks and models. The learning rate and weight decay were set to relatively small values to ensure a steady convergence during training. The batch size was chosen to be as large as possible while fitting into the available memory. We trained the models for $2$ epochs.

We opted to cache the targets of the full-precision model, denoted $f_\theta$. Despite imposing a substantial requirement on disk size --to the tune of several terabytes for each set of the $10K$ samples corresponding to each utilized full-precision model-- this tactic significantly amplified our computational efficiency. Notably, this approach permitted a near doubling of the feasible batch size that could fit in memory.

\label{section:AppendixC}
\section{Ablation Analysis and Extended Discussion}
\label{section:AppendixD}

\subsection{Hybrid Loss for LREC Optimization}
Each experiment was conducted by setting one component of the loss to zero while the other component was retained. This detailed report provides a more granular examination of the significance of the KL divergence and Cross-Entropy components in our INT3 model.

The hyperparameters used for each scenario were varied as follows:

All the experiments were performed with the following fixed parameters: a LLaMA $7$B model quantized with a group size of $128$ at $3$ bits, $r$ set to $32$, with all projections adapted. The learning rate and weight decay were also kept constant at $1 \times 10^{-5}$ in all configurations.

% Now, let us examine the perplexity scores obtained for each scenario on the Wikitext-2 dataset, which serve as performance indicators for the models:

Evidenced in Tables \ref{tab:ablation_study_hyperparams} and \ref{tab:ablation_study_perplexity}, in the absence of Cross-Entropy (CE 0 scenario) the model still retained a remarkably low perplexity of $5.528$. Notably, in this scenario, the loss curves were infinitesimally lagging behind the hybrid loss ones. On the other hand, when KL divergence was omitted (KL 0 scenario), the model showed a slight increase in perplexity to $5.777$, illustrating the role of KL divergence in enhancing prediction capabilities. Nonetheless, the most efficient performance was observed in the Combined scenario, with a perplexity of $5.52$.

These observations underline the combined importance of both components while accentuating the particular role of KL divergence in model stabilization and prediction enhancement. Both components evidently contribute to the model's efficiency, interacting to produce a better result than when acting individually.

\begin{table}[t]
\centering
\begin{tabular}{|l|c|c|c|}
\hline
\textbf{Scenario} & \textbf{CE Coefficient ($\lambda_{ce}$)} & \textbf{KL Coefficient ($\lambda_{kl}$)} & \textbf{Batch Size (bs)} \\
\hline
Cross Entropy 0 & $0.0$ & $1.0$ & $7$ \\
\hline
KL Divergence 0 & $40$ & $0.0$ & $4$ \\
\hline
Combined & $40$ & $1.0$ & $7$ \\
\hline
\end{tabular}
\caption{Hyperparameters configurations for ablation study}
\label{tab:ablation_study_hyperparams}
\end{table}

\begin{table}[t!]
\centering
\begin{tabular}{|l|c|}
\hline
\textbf{Scenario} & \textbf{Perplexity} \\
\hline
Cross Entropy 0 & $5.528$ \\
\hline
KL Divergence 0 & $5.7773$ \\
\hline
Combined & $5.52$ \\
\hline
\end{tabular}
\caption{Perplexity scores on the Wikitext-2 dataset for ablation study}
\label{tab:ablation_study_perplexity}
\end{table}

\subsection{LREC Augmented EMEF}
An intriguing extension to our experimental paradigm was the fusion of our LREC and EMEF methodologies. We first applied LREC to an INT2 quantized $7$B model, successfully error-correcting it to convergence using the parameters specified in Appendix \ref{section:AppendixC}. The resultant model parameters, inclusive of the error-correcting ones, were subsequently frozen. We then proceeded to fine-tune the frozen model through the injection of a novel set of LoRA matrices.

\paragraph{Experimental Parameters} Once again, we utilized the Alpaca dataset for instruction fine-tuning. The micro-batch size and batch size were set at $64$ and $128$ respectively. In order to foster improved model generalization, the instruction inputs during the loss calculation were masked. We selectively injected LoRA weights to the query and key projections, setting the bottleneck parameter $r=8$ and learning rate $lr=3\times 10^{-4}$. The model was fine-tuned over the course of $3$ epochs.

\paragraph{Results and Discussion} The final product of this experiment was a highly competent model, virtually comparable to other Alpaca fine-tuned models in terms of performance. We argue this is compelling evidence that our error-correction strategy is not only effective but also facilitates the fine-tuning of the INT2-quantized model to specialize it. This experiment is particularly groundbreaking as it demonstrates the first instance of a \textbf{fine-tunable, post-training quantized INT2 Large Language Model}. Sample generations emanating from this model are presented in Figure \ref{fig:example_responses}.

\subsection{Effect of the bottleneck parameter \emph{r} and the injection locations on LREC's Performance}

Our study comprised two distinct experiments. First, we evaluated the influence of modifying the bottleneck parameter $r$. Second, we restricted the injection locations solely to the query and key projections. Both alterations resulted in observable instabilities during the training process, which necessitated careful management. Here, we delve into the specifics of these two experiments and their respective outcomes.

\subsection{Influence of Bottleneck Parameter \emph{r}}
Our exploration began by tuning $r$ from its initial value of $32$ down to $24$. Accompanied by a decreased learning rate from $10^{-5}$ to $3 \times 10^{-6}$ and reducing $\lambda_{K L}$ from $1$ to $0.5$, we continued training for the same number of epochs as in the original configuration. This resulted in a model with a perplexity equivalent to the original on the Wikitext dataset ($5.132$), along with comparable training curves. The parameters' reduction demonstrates that it is feasible to achieve an optimum with fewer parameters, akin to the original setup.

As we pushed further to reduce $r$ down to $16$, instabilities started to surface. The training process for a few $4$-bit models suddenly diverged at around $6000-8000$ samples, notwithstanding their initial similarity to the $r=32$ model's trajectory. These models exhibited a slightly higher loss but suggested that these instabilities could be mitigated with careful tuning, potentially leading to models with a minimal number of injected parameters. Case in point, we achieved the same perplexity as the INT4 LREC model of $5.132$ as $r=32$ by decreasing $r$ down to $8$, with $\lambda_{K L}=1$. The balance between optimal performance and model stability remained a crucial consideration throughout these experiments.

\subsection{Restricted Injection Locations}
In addition to tuning $r$, we experimented with limiting injection locations. The specific layers to which we adapted were confined to the query and key projections, limiting the number of layers to just $2$. This strategy was applied to both INT4 and INT2 models.

For the INT2 models, this restriction led to a few instabilities, with the models diverging abruptly after $1$ epoch. The training curves suggest that despite a slower decrease in loss, the models seem to converge slightly higher than their fully injected counterparts. This outcome could indicate that limiting the injection to just two layers might not be sufficient for int2, although further investigations are required to confirm this hypothesis.

In contrast, the INT4 models demonstrated an intriguing behavior. With the same restriction applied, we obtained a perplexity of $5.16$ for $r=8$ and $r=32$, both with a learning rate of $3 \times 10^{-6}$. This result signifies that even with a reduced amount of parameters (bottleneck $r=8$) and injection locations limited to the query and key projections, we could still achieve a performance comparable to the original configuration albeit with a slight degradation.

\subsection{Considerations for Parameter Injection}
From our study, we observe that reducing the number of parameters and limiting the injection locations can still maintain near-optimal performance, albeit with potential instabilities during training. This observation implies that if the goal is to introduce the smallest number of parameters possible, the injection hyperparameters—particularly the bottleneck $r$ and the injection locations—may not be optimal in their current configuration and could potentially be further minimized. However, achieving this requires careful management and tuning of other hyperparameters to stabilize the training process. The pursuit of further reductions and their subsequent impacts on model performance remain open questions for future research.

%\subsection{Additional LREC experiments}
%Finally, to bolster our claims, we provide additional perplexities for the $7$B model in Table \ref{tab:LREC}.

\subsection{Row-wise quantization}
We studied the behavior of our error correction method with
per-row grouped quantized weights, a paradigm conventionally seen as presenting a suboptimal balance between memory usage and performance. Our hypothesis was that by commencing at a higher point on the loss curve, we could still achieve an equivalent minimum via error correction, hence effecting a more favorable memory conservation. However, our empirical findings did not corroborate this hypothesis. Upon examination, we discovered that the training and validation curves of models implementing per-row grouped quantization were analogous to those of $128$-grouped models, albeit exhibiting a noticeable vertical displacement. These observations signify that our proposed approach did not yield the anticipated memory savings.

\end{document}